\documentclass{article}
\usepackage{colm2024_conference}

\usepackage{booktabs}
\usepackage{graphicx}
\usepackage{enumitem}
\usepackage{wrapfig}
\usepackage{algorithm}
\usepackage{algpseudocode}
\usepackage{multicol}

\usepackage{microtype}
\usepackage{amsmath}
\usepackage{colortbl}
\usepackage[utf8]{inputenc}
\definecolor{lightgray}{rgb}{0.9,0.9,0.9}
\usepackage{caption}
\usepackage{subcaption}
\usepackage{setspace}
\usepackage{url}
\usepackage{multirow}
\usepackage{colortbl}
\usepackage{tabularx}
\usepackage{blindtext}
\usepackage{pgfplots}
\pgfplotsset{compat=1.18} 
\usepackage{tikz}
\usetikzlibrary{er,positioning,bayesnet}
\usepackage{makecell}
\usepackage{tipa}
\usepackage{siunitx}
\usepackage{nicefrac}
\usepackage{tocloft}
\usepackage{listings}
\usepackage[raster,skins]{tcolorbox}
\usepackage{xltabular}
\usepackage{adjustbox}
\usepackage{xurl}
\usepackage{setspace}
\usepackage{threeparttable}

\usepackage{lipsum}
\usepackage{arydshln}
\PassOptionsToPackage{table}{xcolor}
\usepackage{varwidth}

\usepackage{amsmath,amsfonts,bm}

\def\eqref#1{equation~\ref{#1}}

\def\1{\bm{1}}

\DeclareMathAlphabet{\mathsfit}{\encodingdefault}{\sfdefault}{m}{sl}
\SetMathAlphabet{\mathsfit}{bold}{\encodingdefault}{\sfdefault}{bx}{n}

\title{Adaptive Deep Reasoning: Triggering Deep Thinking When Needed}

\usepackage{setspace}

\newcommand{\cofirst}{\textsuperscript{†}}

\author{
	\begin{spacing}{1.5}
		Yunhao Wang\cofirst, Yuhao Zhang\cofirst, Tinghao Yu, Can Xu, Feng Zhang, Fengzong Lian \\
		Tencent Hunyuan Team \\
		\{luciuswang, yuhaozzhang, maxwellyu, leocaxu, jayzhang, faxonlian\}@tencent.com \\
	\end{spacing}
}

\begin{document}

\maketitle

\renewcommand{\thefootnote}{†}
\footnotetext{These authors contributed equally to this work.}

\large\begin{center}\textbf{Technical Report}\end{center}
\vspace{2em}

\begin{abstract}
    Large language models (LLMs) have shown impressive capabilities in handling complex tasks through long-chain reasoning. However, the extensive reasoning steps involved can significantly increase computational costs, posing challenges for real-world deployment. Recent efforts have focused on optimizing reasoning efficiency by shortening the Chain-of-Thought (CoT) reasoning processes through various approaches, such as length-aware prompt engineering, supervised fine-tuning on CoT data with variable lengths, and reinforcement learning with length penalties. Although these methods effectively reduce reasoning length, they still necessitate an initial reasoning phase. More recent approaches have attempted to integrate long-chain and short-chain reasoning abilities into a single model, yet they still rely on manual control to toggle between short and long CoT.
    In this work, we propose a novel approach that autonomously switches between short and long reasoning chains based on problem complexity. Our method begins with supervised fine-tuning of the base model to equip both long-chain and short-chain reasoning abilities. We then employ reinforcement learning to further balance long and short CoT generation while maintaining accuracy through two key strategies: first, integrating reinforcement learning with a long-short adaptive group-wise reward strategy to assess prompt complexity and provide corresponding rewards; second, implementing a logit-based reasoning mode switching loss to optimize the model’s initial token choice, thereby guiding the selection of the reasoning type.
    Evaluations on mathematical datasets demonstrate that our model can dynamically switch between long-chain and short-chain reasoning modes without substantially sacrificing performance. This advancement enhances the practicality of reasoning in large language models for real-world applications.
\end{abstract}

\section{Introduction}
\label{sec:intro}
Large Language Models (LLMs) have exhibited significant capabilities in handling complex tasks. Notably, recently proposed models such as OpenAI's o-series models \cite{openai2024openaio1card}, DeepseekR1 \cite{deepseekai2025deepseekr1incentivizingreasoningcapability}, and QwQ \cite{qwq32b} have demonstrated substantial performance improvements by emulating human-like problem-solving processes. These models enhance LLMs with long-thought reasoning abilities, enabling them to decompose complex problems into sub-tasks, iteratively identify and correct errors, simplify intricate steps, and explore alternative strategies when initial approaches are insufficient.
While the process of generating answers through thoughtful reasoning can activate the reasoning capabilities of LLMs, it often results in intermediate reasoning steps that are significantly longer than the desired final answers. This increased length can elevate inference costs and impede the application of these models in various real-world scenarios.

Recent research has concentrated on minimizing the propensity of reasoning models to overthink and improving reasoning efficiency. Various strategies have been proposed to achieve this goal. One approach involves controlling the length of reasoning through input prompt engineering, such as setting a token budget in prompts to reduce excessive reasoning tokens \cite{han2025tokenbudgetawarellmreasoning}, or adjusting prompts to restrict computations to single steps \cite{chen2024unlockingcapabilitiesthoughtreasoning}. Another method employs supervised fine-tuning with variable-length Chain-of-Thought (CoT) data to foster more concise reasoning. Techniques like TokenSkip \cite{xia2025tokenskipcontrollablechainofthoughtcompression} and C3oT \cite{kang2024c3otgeneratingshorterchainofthought} aim to produce shorter reasoning chains by either omitting irrelevant steps or compressing the reasoning process. Additionally, reinforcement learning with length-based rewards has been utilized to optimize reasoning by incentivizing shorter, correct answers and penalizing longer or incorrect ones. Methods such as O1-Pruner \cite{luo2025o1prunerlengthharmonizingfinetuningo1like} and Demystifying-Long-CoT \cite{yeo2025demystifyinglongchainofthoughtreasoning} use rewards to reduce CoT length, effectively mitigating overthinking without compromising accuracy.

However, although these methods can reduce CoT length, they still necessitate an initial "thinking" phase before producing an answer, requiring users to wait for the reasoning process regardless of the question type. Recently, models such as Claude 3.7 Sonnet \cite{claude-3-7-sonnet}, Qwen3 \cite{yang2025qwen3technicalreport} and Llama-Nemotron \cite{bercovich2025llamanemotronefficientreasoningmodels} have been developed to generate both short and long CoT using the same model with different control prompts. Nevertheless, these models cannot autonomously determine which type of CoT to employ.

In this paper, we present a novel method that automatically selects between short and long reasoning chains when addressing new problems, thereby optimizing the reasoning length for each specific problem. Our approach begins with training a supervised fine-tuning model equipped with both short and long chain-of-thought reasoning capabilities by providing it with data for both reasoning types. We then enhance the SFT model using reinforcement learning to balance these reasoning abilities, allowing it to generate short CoT when appropriate without impacting overall performance.
We employ two strategies to achieve this balance. The first strategy integrates reinforcement learning with a group-wise reward strategy, which assesses the correctness of all sampled responses to a given prompt to determine its complexity. This assessment guides the selection of the appropriate reasoning type and corresponding rewards. 
The second strategy involves implementing a logits-based reasoning mode switching loss on the first generated token, which determines the reasoning type of the response. The suitable reasoning type for each prompt is derived from the overall correctness of all sampled responses during the group-wise rewarding phase. This approach aids the model in learning to discern whether to produce long-chain or short-chain reasoning.
We evaluate our model's performance on multiple mathematical capability datasets with varying difficulty levels. Experimental results demonstrate that our model can autonomously choose between short and long CoT without compromising efficiency, effectively reducing the average response length.

\section{Related Work} 
Utilizing long-chain reasoning, which involves extensive reasoning processes before arriving at a final answer, has proven effective for addressing complex tasks \cite{openai2024openaio1card, deepseekai2025deepseekr1incentivizingreasoningcapability, qwq32b}. However, excessively lengthy reasoning chains can result in significant computational overhead and delay the generation of final answers, rendering them sub-optimal for many scenarios.

Consequently, many studies have concentrated on efficient reasoning strategies, particularly aimed at minimizing the tendency of reasoning models to overthink. One direct method to control reasoning length is through input prompt engineering. Techniques include setting a token budget in prompts to restrict excessive reasoning tokens \cite{han2025tokenbudgetawarellmreasoning}, or modifying prompts to reduce computations of single steps \cite{chen2024unlockingcapabilitiesthoughtreasoning}. However, the effectiveness of these strategies relies on domain-specific tuning and may not be universally applicable.

Another approach to model-based efficient reasoning involves fine-tuning large language models with variable-length chain-of-thought data to enhance reasoning efficiency. This typically involves creating datasets with varying lengths of reasoning steps and applying supervised fine-tuning to enable LLMs to learn more compact reasoning chains. Techniques such as TokenSkip \cite{xia2025tokenskipcontrollablechainofthoughtcompression} and C3oT \cite{kang2024c3otgeneratingshorterchainofthought} focus on generating shorter CoT data by skipping or compressing reasoning steps. TokenSkip assesses the semantic importance of each reasoning component and reduces unnecessary tokens, while C3oT employs GPT-4 as a compressor to ensure that compressed reasoning retains all essential information.

Some research utilizes reinforcement learning integrated with length-based rewards to improve the efficiency of reasoning processes. This approach aims to shorten the reasoning process by assigning higher rewards to shorter, correct answers while penalizing lengthy or incorrect ones. Studies such as \cite{luo2025o1prunerlengthharmonizingfinetuningo1like, yeo2025demystifyinglongchainofthoughtreasoning,kimiteam2025kimik15scalingreinforcement} have explored various reinforcement learning techniques to optimize reasoning length. For example, O1-Pruner \cite{luo2025o1prunerlengthharmonizingfinetuningo1like} introduces a Length-Harmonizing Reward combined with a PPO-style loss to effectively reduce the chain-of-thought length. Similarly, Demystifying-Long-CoT \cite{yeo2025demystifyinglongchainofthoughtreasoning} uses a Cosine Reward to control the length of CoT reasoning, demonstrating that reinforcement learning can mitigate the overthinking phenomenon without compromising reasoning accuracy.

Recently, instead of merely shortening long-chain reasoning processes, some works \cite{claude-3-7-sonnet, yang2025qwen3technicalreport, bercovich2025llamanemotronefficientreasoningmodels} have enabled models to flexibly switch between short-form responses and long-chain reasoning as needed through prompt-based control. This approach combines the efficiency of short-chain reasoning with the comprehensive capabilities of long-chain reasoning within a single model. However, these methods typically require manual control to toggle the reasoning mode on or off.

Our method improves upon existing approaches by automatically switching between long-chain and short-chain reasoning forms based on the complexity of the task, thereby removing the need for manual intervention or specific prompts. This automatic adjustment ensures that the reasoning length is optimally tailored to each problem, enhancing the overall user experience.


\section{Method}
\label{sec:approch}

In this study, we present a novel approach that autonomously transitions between short and long reasoning chains according to problem complexity, utilizing supervised fine-tuning and reinforcement learning. In Section \ref{sft}, we describe the process of using supervised fine-tuning to endow base models with both long-chain and short-chain reasoning capabilities. Section \ref{rl} presents the application of reinforcement learning to further balance the generation of short and long chains of thought through a long-short adaptive group-wise reward strategy. In Section \ref{rmsl}, we introduce the reasoning mode switching loss, which aids in guiding the selection of the appropriate reasoning type.

\subsection{Supervised Fine-Tuning for Adaptive Reasoning} \label{sft}

To equip the base model with both short and long chain-of-thought reasoning capabilities, we employ both short and long chain-of-thought data for supervised fine-tuning. Additionally, for each reasoning mode, we employ both instructional and non-instructional data. This allows the SFT model to explicitly control the generation of both long and short chains of thought during reinforcement learning sampling through instruction. Although conventional prefix decoding (e.g., prepending "\textless think\textgreater") can be used to select the reasoning mode, it often compromises proficiency. In contrast, explicit instruction helps the model preserve its reasoning capacity for generating both long and short reasoning chains. The supervised fine-tuning dataset, which includes four training categories, is outlined in Table \ref{tab:cot_types}.

\begin{table}[ht]
	\centering
	\begin{tabular}{@{}l l l c@{}}
		\toprule
		\textbf{Index} & \textbf{Mode} & \textbf{Instruction} & \textbf{Example} \\ 
		\midrule
		1 & Long CoT & w/ instruction & [Please answer with Long CoT] + Query \\
		2 & Long CoT & w/o instruction & Query \\
		\midrule
		3 & Short CoT & w/ instruction & [Please answer with Short CoT] + Query \\
		4 & Short CoT & w/o instruction & Query \\
		\bottomrule
	\end{tabular}
        \caption{Four training categories of supervised fine-tuning dataset}
        \label{tab:cot_types}
\end{table}
Through this hybrid training approach, the SFT model gains fundamental adaptive reasoning abilities. Additionally, it facilitates explicit instruction-based control over long and short reasoning modes, while preserving a high level of proficiency.

\subsection{Reinforcement Learning for Adaptive Reasoning} \label{rl}

Following supervised fine-tuning, we employ reinforcement learning (RL) to further improve the model's ability to adaptively reason over both long and short contexts. This approach enables the model to select suitable reasoning strategies based on the complexity of the problem. We utilize the Group Relative Policy Optimization (GRPO) method \citep{shao2024deepseekmath} for RL. This method provides improved optimization efficiency compared to the traditional Proximal Policy Optimization (PPO).

\subsubsection{Reasoning Mode Controlled Sampling}
During GRPO sampling, we augment user prompts with explicit  instructions to force model to generate long cot and short cot  samples with equal num. The instructions are identical to those utilized during SFT, as illustrated in Table \ref{tab:cot_types}. Specifically, for each question \( q \), the previous policy model \( \pi_{\text{old}} \) generates a set of responses \( \mathcal{G} \) in the following manner:
\begin{equation}
	\mathcal{G} = \underbrace{\{o^{\text{long}}_1, \ldots, o^{\text{long}}_{G/2}\}}_{\text{Long CoT}} \cup \underbrace{\{o^{\text{short}}_{G/2+1}, \ldots, o^{\text{short}}_G\}}_{\text{Short CoT}}
\end{equation}
This sampling method ensures an equal number of samples for each prompt, facilitating the evaluation of which mode—long or short reasoning—is more suitable for each prompt by considering the accuracy of both modes. This approach allows the model to learn from both reasoning modes and to switch between them more effectively.

\subsubsection{Reward Modeling}

The reward model serves as environmental feedback in reinforcement We employ a hybrid reward system that integrates both LLM-based and rule-based components to provide reward signals. This system comprises two key elements:

\begin{itemize}
    \item \textbf{Correctness Reward}: We utilize a 7-billion parameter LLM to evaluate the consistency between the model's output and the standard answer. This evaluation is conducted in a step-by-step manner, taking into account the type of question, the presence of multiple sub-questions, and the accuracy of the final result. Compared to rule-based rewards, this approach exhibits greater accuracy when handling complex formulas and specific formatting requirements in responses.
    \item \textbf{Format Reward}: For extended reasoning chains, the model verifies whether the output includes reasoning-related special tokens (e.g., \textless think\textgreater, \textless answer\textgreater) and whether the format adheres to predefined standards.
\end{itemize}

The aforementioned reward system provides fundamental signals that reflect the quality of each sampling. Based on these signals, we further perform reward shaping, which assists the model in choosing between long and short reasoning modes during reinforcement learning training. 

\subsubsection{Reward Shaping} \label{reward_shaping}
To enable the model to automatically choose between long and short reasoning modes and to mitigate the tendency for overthinking in long reasoning chains, we propose a problem complexity based adaptive reward strategy. This strategy assigns rewards by jointly considering the complexity of the problem and the corresponding reasoning mode. Additionally, we implement a reward warm-up method to further stabilize the reinforcement learning training process and apply a length penalty to reduce redundancy in long-chain reasoning.

    \textbf{Long-short Adaptive Group-wise Reward}: 
    Given that long-chain reasoning is effective for addressing complex problems, while short-chain reasoning can yield good results for relatively easy problems, we aim for our model to switch between long and short reasoning modes based on the complexity of the problem. This approach allows the model to leverage the advantages of both reasoning modes. Instead of determining problem difficulty through conventional metadata-based methods, such as classification by discipline or grade level, we estimate it based on the sampling accuracy of each prompt. Using this estimation, we design adaptive rewards that assign different values to each sample according to the correctness of the answer and the complexity of the problem. 

    During response sampling, incorrect samples incur a penalty of -1. When the accuracy of short CoT responses to a prompt \( p \) surpasses a specified threshold \(\theta\), it suggests that the problem is relatively easy, allowing the model to effectively solve it using short CoT. Consequently, correct short CoT responses are awarded a higher reward of +1.5, compared to correct long CoT responses, which receive a reward of +1.0. Conversely, if the accuracy of short CoT responses falls below the threshold \(\theta\), indicating a more complex problem, correct long CoT responses are granted a higher reward than correct short CoT responses.
    The reward shaping scheme is as follows:

	\begin{equation}
		\mathcal{R}(p,\alpha, \theta) = 
		\begin{cases}
			-1.0 & \text{Incorrect Answer} \\[6pt]
			\begin{aligned}
				\text{Case } \alpha > \theta: &\quad
				\begin{cases} 
					+1.5 & \text{(Short CoT)} \\ 
					+1.0 & \text{(Long CoT)}
				\end{cases} \\[6pt]
				\text{Case } \alpha \leq \theta: &\quad
				\begin{cases} 
					+1.0 & \text{(Short CoT)} \\ 
					+1.5 & \text{(Long CoT)}
				\end{cases}
			\end{aligned}
			& \text{Correct Answer}
		\end{cases}
	\end{equation}
	
	where:
	\begin{itemize}
            \item $p$: Prompt for sampling 
		\item $\alpha \in [0,1]$: Accuracy of short CoT samples for prompt $p$
		\item $\theta \in [0,1]$: Accuracy threshold for short CoT
	\end{itemize}

	 \textbf{Reward Warmup}: 
	Our reward shaping strategy dynamically assigns different rewards to long chain reasoning and short chain reasoning based on the complexity of the problem. During the early stages of reinforcement learning  training, when the model's preference for long or short chains is still unstable, we implement a reward warm-up mechanism to prevent one type of reasoning from dominating and causing training collapse. Specifically, at the beginning of our model training, short chain reasoning is predominant. Therefore, we apply a warm-up to the short chain rewards. We gradually increase the short chain reward from being equal to the long chain reward to the target value when the accuracy of short CoT samples exceeds the threshold \(\theta\). The warm-up schedule is as follows:

        \begin{equation}
		\mathcal{R}_{correct\_answer}(p,\alpha, \theta) = \begin{cases} 
			\displaystyle \frac{T_{\text{current\_step}}}{T_{\text{warmup}}} \times 0.5 + 1 & (T_{\text{current\_step}} < T_{\text{warmup}}) \\
			1 & (T_{\text{current\_step}} \geq T_{\text{warmup}})
		\end{cases}
	\end{equation}
	\textbf{Soft Length Penalty}: To prevent excessive elaboration in long chain reasoning, we incorporate a length penalty into the reward system for long chains of thought. When multiple reasoning paths lead to the correct answer, shorter traces are granted higher rewards, thus reducing redundancy while maintaining accuracy. We adopt the length penalty from the DAPO \citep{yu2025dapo}, but in our approach, we define the penalty interval $L_{\Delta}$ as the difference between the longest reasoning chain and the average length of short chains, which penalize longer responses within a specified interval as follows:

        \begin{equation}
        \mathcal{R}_{{length\_penalty}}(y) = 
        \begin{cases} 
        0 & \text{if $y$ is short cot, or }  |y| \leq L_{\max} - L_{\Delta}, \\
        \frac{(L_{\max} - L_{\Delta}) - |y|}{L_{\Delta}} & \text{if } L_{\max} - L_{\Delta} < |y| \leq L_{\max}, \\
        -1 & \text{if } |y| > L_{\max}.
        \end{cases}
        \end{equation}
	where:
	\begin{itemize}
		\item $y$: sampled response
		\item $L_{\text{max}}$: the maximum length of responses for a given prompt.
		\item $L_{\Delta}$: $L_{\Delta} = L_{\text{max}} - {L}_{\text{min}}$, where ${L}_{\text{min}}$ represents the average length of short reasoning responses.
	\end{itemize}
    This penalty is applied in conjunction with the problem complexity-based adaptive reward.

\subsection{Reasoning Mode Switching Loss} \label{rmsl}
The reasoning style of a model primarily depends on the first response token (e.g., the initial token of "\textless think\textgreater"). Therefore, learning the probability distribution of this first generated token is crucial for the model to effectively switch between long-chain and short-chain reasoning modes. However, since reinforcement learning algorithms optimize the generation distribution of all tokens in a response simultaneously, the weight assigned to the first token is relatively low, particularly in long-chain reasoning modes. This often results in reasoning mode switching under-optimized.

To address this issue, we propose a reasoning mode switching loss applied to the first generated token in the response, which compels the model to explicitly learn how to switch reasoning modes. We begin by categorizing the softmax-normalized logits for generation into two groups. The first group, denoted as \(\ell_{\text{long}}\), includes the logits of tokens that would initiate a long-chain reasoning mode. The second group comprises the top-k logits of tokens, excluding those in the first group, and is referred to as \(\ell_{\text{short}}\). We then construct the reasoning mode switching loss using a margin-ranking loss:
\begin{equation}
L_{rmsl} = 
\begin{cases} 
	\max(0, \sum \ell_{\text{long}}\ - \sum \ell_{\text{short}}\ + \text{margin}_1) & \text{if } \alpha \ge \theta \\
	\max(0, \sum \ell_{\text{short}}\ - \sum \ell_{\text{long}}\ + \text{margin}_2) & \text{if } \alpha < \theta
\end{cases}
\end{equation}
where $\alpha$ denotes the accuracy of short-chain responses sampled during reinforcement learning for a given prompt, and $\theta$ represents the accuracy threshold for short CoT responses, as discussed in Section \ref{reward_shaping}.

By incorporating the GRPO loss and the reasoning mode switching loss, we arrive at our final optimization objective:
\begin{equation}
L = L_{grpo} + \lambda \times L_{rmsl} 
\end{equation}
Here, \(\lambda\) is a coefficient that balances the contributions of \(L_{grpo}\) and \(L_{rmsl}\).

\section{Experiment}
\label{sec:experiment}

\subsection{Benchmarks}
We evaluate our model on a range of mathematical benchmarks to assess its long-chain and short-chain reasoning abilities, including MATH-500 \cite{lightman2023letsverifystepstep}, AIME-100, AIME 2024 \cite{aops2024aimeii1}, AIME 2025 \cite{aops2025aimeii1}, Super GPQA Math \cite{pteam2025supergpqascalingllmevaluation}, and Olympiad Bench Math \cite{he2024olympiadbenchchallengingbenchmarkpromoting}. Since AIME 2024 and AIME 2025 each contain only 60 problems, we constructed the AIME-100 benchmark by randomly sampling 100 problems from historical AIME exam questions.

\subsection{Experimental Setup}

\textbf{Training Data}: We gathered a collection of mathematical problems from junior and senior high school competitions, as well as from Olympiad-level challenges. To ensure high data quality, we exclude proof-based questions, multiple-choice items, and true/false problems from the dataset.
We employed rejection sampling to collect both long and short CoT data using state-of-the-art reasoning models. To ensure the accuracy of this data, we used an evaluation LLM, which have been utilized in reward modeling, to assess the consistency between the model's output and the standard answers. Ultimately, we compiled an SFT dataset comprising 220k long chain-of-thought  reasoning entries and 220k short CoT reasoning entries, as well as an RL dataset consisting of 20k problems.

\textbf{Base Model}: We use Qwen-2.5-Math-Instruct-7B \cite{yang2024qwen25mathtechnicalreportmathematical} as our base model, which is specifically designed for mathematical reasoning tasks.

\textbf{Training Configuration}: For surpervised fine-tuning, we train the base model using a 32k token context window and a global batch size of 64. The learning rate is set to 6e-5, and we employ a cosine learning rate schedule that decays to 6e-6 over the course of 2 epochs. 
For reinforcement learning, we employed a batch size of 64 and 16 trajectory rollouts. The learning rate was set to 5e-7, and the training was conducted for one epoch.
The training is conducted on 64 Nvidia H20 GPUs.

\subsection{SFT Model Evaluation}
We employ a test set that includes both primary-level and competitive-level problems to evaluate the performance of the supervised fine-tuned models. The experimental results are presented in Table \ref{tab:sft_evaluation}. These results demonstrate that training with mixed SFT data preserves both long chain and short chain reasoning abilities, compared to training with only one type of reasoning data. This approach provides a solid foundation for enabling the model to automatically switch between these two reasoning modes.

\begin{table}[!ht]
  \centering
  \small 
  \renewcommand{\arraystretch}{1.4} 
  \setlength{\tabcolsep}{0.4em}  
  
  \begin{tabular}{lcc} 
    \toprule[1pt] 
    \textbf{Model} & \textbf{Accuracy} & \textbf{Thinking Rate} \\
    \midrule
    Long-chain only SFT & 58.0\% & 100\% \\
    Short-chain only SFT & 44.8\% & 0\% \\
    Mixed-SFT with long CoT instruction & 57.1\% & 100\% \\
    Mixed-SFT with short CoT instruction & 42.4\% & 0\% \\
    Mixed-SFT without instruction & 58.0\% & 92.9\% \\
    \bottomrule[1pt] 
  \end{tabular}
  
  \vspace*{0.5\baselineskip} 
  \caption{Comparison of SFT Models Employing Various Training and Decoding Strategies}
  \label{tab:sft_evaluation}
\end{table}

\subsection{RL Model Evaluation}

The results of the reinforcement learning experiments are presented in Table \ref{tab:model_comparison}. Our models—RL-AR-Exp1 and RL-AR-Exp2, which are trained with long-short adaptive group-wise reward, and RL-AR-RMSL, which incorporates an additional ranking model switching loss—demonstrate the ability to effectively switch between long and short reasoning modes in response to the complexity of the data. For instance, on relatively easy datasets such as MATH-500, all three models maintain a low proportion of long-chain reasoning while still achieving high accuracy. In contrast, for more challenging datasets such as AIME-2024 and AIME-2025, which consist solely of difficult problems, the models predominantly employ long-chain reasoning to solve the problems, thereby achieving accuracy comparable to that of models dedicated to long-chain reasoning.

{
\renewcommand{\arraystretch}{1.3}

\begin{table*}
        \scriptsize
	\centering
        \setlength{\tabcolsep}{10pt}
	\begin{tabular}{cccccccc cccc}
		\toprule
		
            &\textbf{SFT-Short} &\textbf{SFT-Long} &\textbf{RL-AR-Exp1} &\textbf{RL-AR-Exp2} &\textbf{RL-AR-RMSL} \\
		
            \midrule
		MATH-500  & 80.0/0\%/731 & \textbf{91.2}/100\%/3631 & 83.8/22\%/2116 & 87.6/38\%/2437 & \underline{89.2}/39\%/2486 \\
            SuperGPQA-Math  & 32.0/0\%/2622 & 53.8/100\%/9571 & 44.3/56\%/6962 & \underline{53.9}/93\%/8663 & \textbf{54.1}/95\%/8976  \\
		OlympiadBench-Math  & 55.4/0\%/2329 & \textbf{76.1}/100\%/6012 & 67.5/46\%/4285 & 73.1/72\%/5095 & \underline{74.6}/81\%/5396 \\
		AIME-100  & 39.0/0\%/1502 & \underline{59.0}/100\%/10264 & 55.0/70\%/7968 & \textbf{64.0}/88\%/9041 &  58.0/76\%/8132 \\ 
		AIME-2024  & 13.3/0\%/1599 & \underline{56.7}/100\%/12047 & 53.3/100\%/12168 & 53.3/100\%/12647 & \textbf{63.3}/100\%/12472 \\
		AIME-2025  & 6.7/0\%/2009 & \textbf{36.7}/100\%/14672 & \underline{26.7}/86\%/12366 & \textbf{36.7}/100\%/12917 & \textbf{36.7}/100\%/13047 \\
		\bottomrule
	\end{tabular}
	\caption{Model performance on various mathematical benchmarks. Each column reports three metrics: accuracy, the proportion of long-chain reasoning, and the average number of response tokens. The highest values for each metric are highlighted in bold, while the second-highest values are underlined.}
	\label{tab:model_comparison}
\end{table*}
}

\section{Conclusion}
\label{sec:conclusion}

In this study, we propose a method that enables a reasoning model to dynamically switch between short and long reasoning chains depending on the complexity of the problem. Initially, we enhance the base model with both long and short chain reasoning capabilities through supervised fine-tuning. Subsequently, we employ reinforcement learning, utilizing group-wise rewards and logit-based ranking margin loss, to train the model to determine the appropriate reasoning mode for new problems. This approach allows our model to perform efficient reasoning without significantly affecting its performance. Evaluations on mathematical tasks demonstrate the model's ability to adaptively switch reasoning modes. This advancement facilitates more efficient reasoning for large reasoning models.

\clearpage
\bibliography{colm2024_conference}
\bibliographystyle{colm2024_conference}

\end{document}